\documentclass[11pt,letterpaper]{article}
\usepackage{emnlp2017}
\usepackage{times}
\usepackage{latexsym}

\usepackage{url}
\usepackage{amsmath}
\usepackage{amsthm}
\usepackage{amssymb}
\usepackage{amsfonts,eucal,amsbsy}
\usepackage{mathtools}
\usepackage{booktabs}
\usepackage{caption}
\usepackage{subcaption}
\usepackage{multirow}
\usepackage{wrapfig}
\usepackage{xcolor, colortbl}
\usepackage{fixltx2e}
\usepackage{pbox}
\usepackage[titletoc,title]{appendix}

\emnlpfinalcopy

\def\emnlppaperid{938}

\definecolor{navyblue}{rgb}{0.03, 0.57, 0.82}
\definecolor{yellow}{rgb}{1.0, 0.96, 0.0}

\newcommand{\cellattn}[2]{\cellcolor{navyblue!#1}{#2}}

\title{Reference-Aware Language Models}

\author{Zichao Yang$^1$\thanks{Work completed at DeepMind.} , Phil Blunsom$^{2,3}$, Chris Dyer$^{1,2}$, and Wang Ling$^2$ \\
  $^1$Carnegie Mellon University, $^2$DeepMind, and $^3$University of Oxford\\
  {\tt zichaoy@cs.cmu.edu, \{pblunsom,cdyer,lingwang\}@google.com} }

\date{}

\hypersetup{draft}

\begin{document}

\maketitle

\begin{abstract}
  We propose a general class of language models that treat reference as discrete
  stochastic latent variables. This decision allows for the creation of entity 
  mentions by accessing external databases of referents (required by, e.g., dialogue generation) or
  past internal state (required to explicitly model coreferentiality). Beyond 
  simple copying, our coreference model can additionally refer to a referent using varied 
  mention forms (e.g., a reference to ``Jane'' can be realized as ``she''), a characteristic 
  feature of reference in natural languages. Experiments on three
  representative applications show our model variants outperform models based
  on deterministic attention and standard language modeling baselines.
\end{abstract}

\section{Introduction}
Referring expressions (REs) in natural language are noun phrases (proper nouns,
common nouns, and pronouns) that identify objects, entities, and events in an
environment. REs occur frequently and they play a key role in communicating
information efficiently. While REs are common in natural language, most
previous work does not model them explicitly, either treating REs as ordinary
words in the model or replacing them with special tokens that are filled in with
a post processing step \cite{wen:2016,LuongSLVZ15}. Here we propose a language modeling
framework that explicitly incorporates reference decisions. In part, this is based on 
the principle of pointer networks in which copies are made from another source \citep{GulcehreANZB16,GuLLL16,ling:2016,ptrnets,ahn:2016, merity2016pointer}. 
However, in the full version of our model, we go beyond simple copying and enable 
coreferent mentions to have different forms, a key characteristic of natural language reference.

\begin{figure}[!tb]
  \centering
  \includegraphics[width=0.5\textwidth]{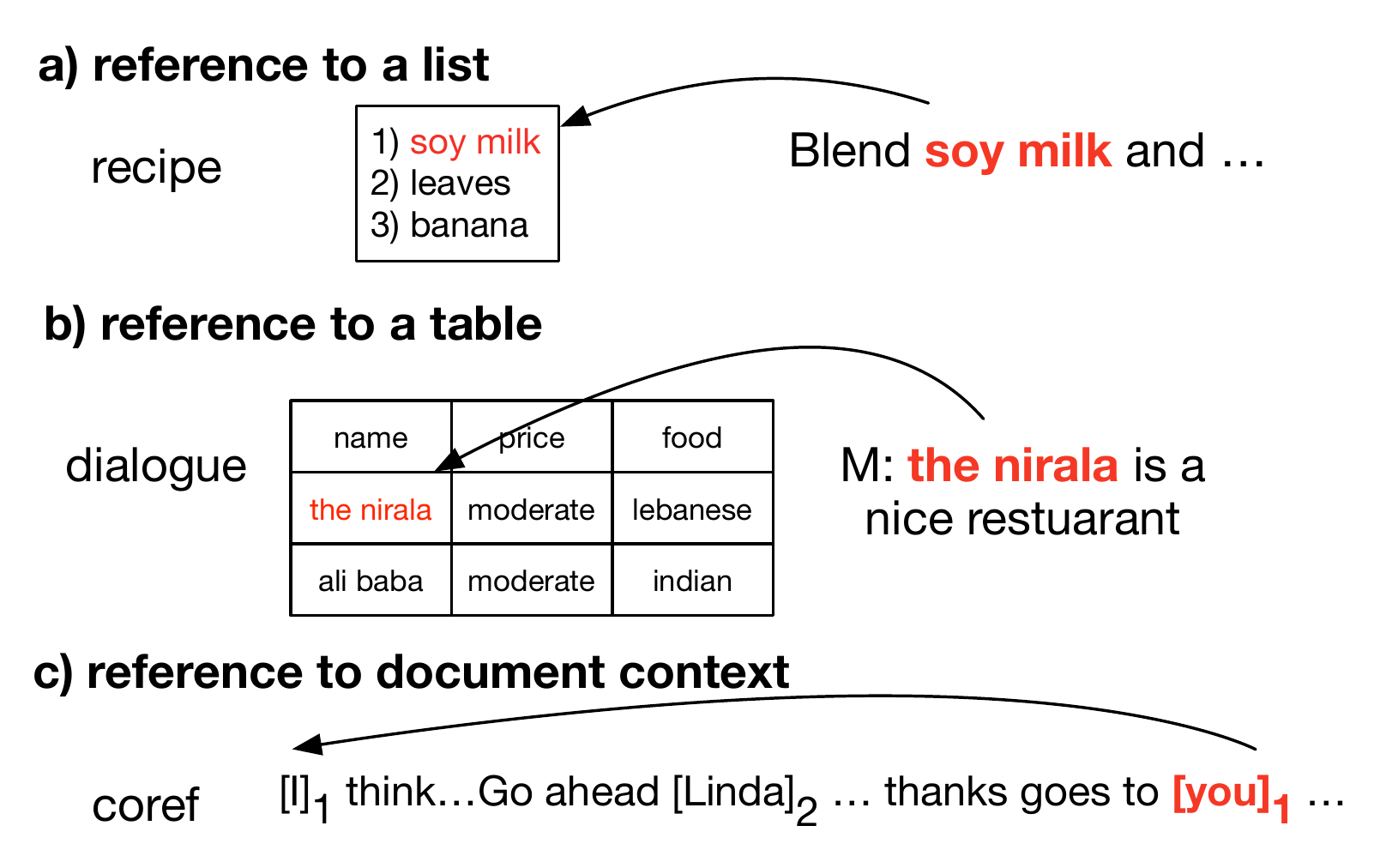}
  \caption{Reference-aware language models.}
  \label{fig:reference}
\end{figure}

Figure~\ref{fig:reference} depicts examples of REs in the context of the three
tasks that we consider in this work.  First, many models need to refer to a list of
items~\citep{kiddon:2016, wen:2016}. In the task of recipe generation from a list of 
ingredients~\citep{kiddon:2016}, 
the generation of the recipe will frequently refer to these items. As shown in
Figure~\ref{fig:reference}, in the recipe ``Blend {\tt soy milk} and \ldots'',
{\tt soy milk} refers to the ingredient summaries. Second, reference to a database is crucial
in many applications. One example is in task oriented dialogue where access to
a database is necessary to answer a user's query~\citep{young2013pomdp,
  li:2016, vinyals:2015, wen:2016, sordoni:2015, serban2016building,
  bordes2016learning, williams2016end, shang2015neural, wen2016network}. Here
we consider the domain of restaurant recommendation where a system refers to
restaurants (name) and their attributes (address, phone number etc) in its
responses. When the system says ``{\tt the nirala} is a nice restaurant'', it
refers to the restaurant name {\tt the nirala} from the database. 
Finally, we address references within a document~\citep{mikolov2010recurrent, ji2015document,
  wang2015larger}, as the generation of words will often refer to previously
generated words.  For instance the same entity will often be referred to
throughout a document.  In Figure~\ref{fig:reference}, the entity {\tt you}
refers to {\tt I} in a previous utterance. In this case, copying is 
insufficient-- although the referent is the same, the form of the mention is different.

In this work we develop a language model that has a specific module for
generating REs. A series of decisions (should I generate a RE? If yes, which
entity in the context should I refer to? How should the RE be rendered?)
augment a traditional recurrent neural network language model and the two
components are combined as a mixture model. Selecting an entity in context is
similar to familiar models of
attention~\citep{BahdanauCB14}, but rather than being a soft
decision that reweights representations of elements in the context, it is
treated as a hard decision over contextual elements which are stochastically
selected and then copied or, if the task warrants it, transformed (e.g., a
pronoun rather than a proper name is produced as output).
% Two variants are
% possible for updating the RNN state: one that only looks at the generated
% output form; and a second that looks at values of the latent variables. The
% former admits trivial unsupervised learning, latent decisions are conditionally
% independent of each other given observed context, whereas the latter enables
% more expressive models that can extract information from the entity that is
% being referred to.
In cases when the stochastic decision is not available in training, we treat it
as a latent variable and marginalize it out. For each of the three tasks, we
pick one representative application and demonstrate our reference aware model's
efficacy in evaluations against models that do not explicitly include a
reference operation.

Our contributions are as follows:
\begin{itemize}
\item We propose a general framework to model reference in language. We
  consider reference to entries in lists, tables, and document context. We
  instantiate these tasks into three specific applications: recipe generation,
  dialogue modeling, and coreference based language models.
\item We develop the first neural model of reference that goes being 
copying and can model (conditional on context) how to form the mention.
\item We perform comprehensive evaluation of our models on the three data sets
  and verify our proposed models perform better than strong baselines.
\end{itemize}

%We select three applications that are representative to show reference-aware
%language models can be widely applied: modeling dialogue (here, the REs can be
%copied from a database), modeling the generation of recipes given a set of
%ingredients (here, the REs refer to ingredients), and modeling language with
%explicit coreference decisions (here, the referents are previous REs in the
%text stream; and, rather than just copying, new words, like pronouns, can be
%produced).

\section{Reference-aware language models}
Here we propose a general framework for reference-aware language models.

We denote each document as a series of tokens $x_1,\ldots, x_L$, where $L$ is
the number of tokens.  Our goal is to maximize $p(x_i \mid c_i)$, the
probability of each word $x_{i}$ given its previous context
$c_i = x_1, \ldots ,x_{i-1}$.  In contrast to traditional neural language
models, we introduce a variable $z_i$ at each position, which controls the
decision on which source $x_i$ is generated from. Then the conditional
probability is given by:
\begin{align}
  p(x_i, z_i \mid c_i) = p(x_i \mid z_i, c_i) p (z_i \mid c_i), \label{eq:marg}
\end{align}
where $z_{i}$ has different meanings in different contexts. If $x_{i}$ is from
a reference list or a database, then $z_{i}$ is one dimensional and $z_{i		} = 1$
denotes $x_{i}$ is generated as a reference. $z_i$ can also model more complex
decisions. In coreference based language model, $z_{i}$ denotes a series of
sequential decisions, such as whether $x_{i}$ is an entity, if yes, which
entity $x_{i}$ refers to. When $z_i$ is not observed, we will train our model
to maximize the marginal probability over $z_{i}$, i.e., 
$p(x_i | c_i) = \sum_{z_i}p(x_i|z_i, c_i) p(z_i|c_i)$.

\subsection{Reference to lists}
\begin{table*}[!thbp]
  \centering
  \begin{tabular}{l | l}
    Ingredients & Recipe \\
    \hline
    1 cup plain \texttt{soy milk} & \multirow{3}{7cm}{Blend \texttt{soy milk} and \texttt{spinach leaves} together in a blender until smooth. Add \texttt{banana} and pulse until thoroughly blended.} \\
    3/4 cup packed fresh \texttt{spinach leaves}  \\
    1 large \texttt{banana}, sliced
  \end{tabular}
  \caption{Ingredients and recipe for {\em Spinach and Banana Power Smoothie}.}
  \label{tab:recipe_example}
\end{table*}

We begin to instantiate the framework by considering reference to a list of items.
Referring to a list of items has broad applications, such as generating documents based
on summaries etc. Here we specifically consider the application of recipe generation 
conditioning on the ingredient lists.
Table.~\ref{tab:recipe_example} illustrates the ingredient list and recipe for
{\em Spinach and Banana Power Smoothie}. We can see that the ingredients
\texttt{soy milk, spinach leaves, and banana} occur in the recipe.
\begin{figure}[!h]
  \centering
  \includegraphics[width=0.4\textwidth]{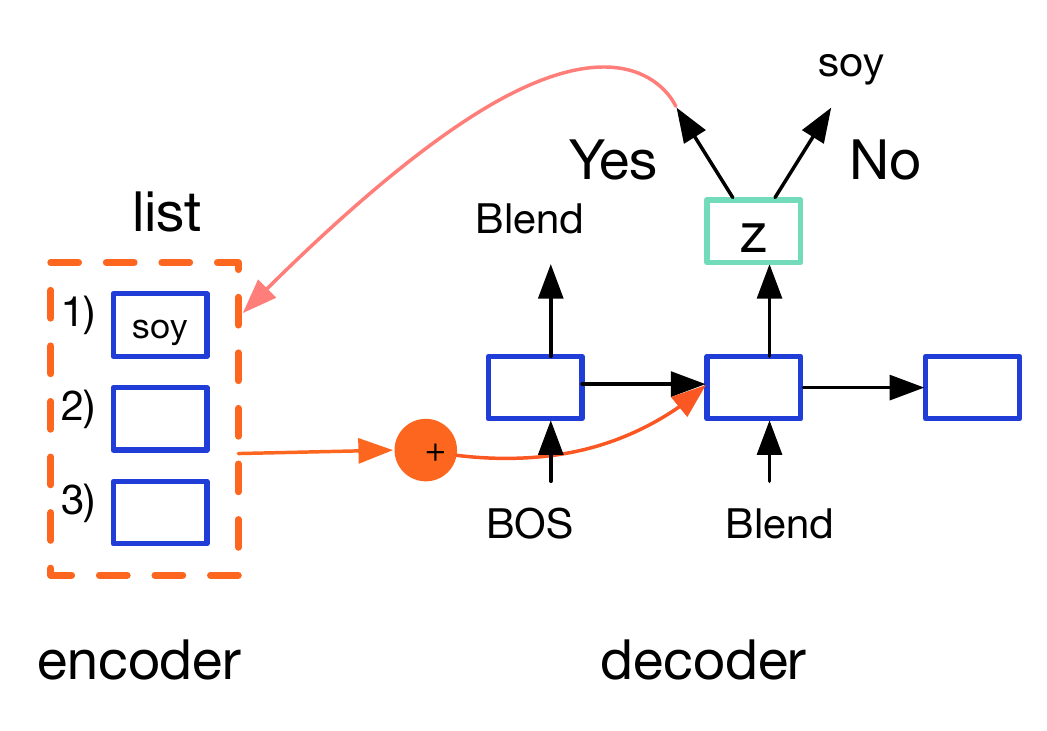}
  \caption{Recipe pointer}
  \label{fig:recipe_pointer}
\end{figure}

Let the ingredients of a recipe be $X = \{x_{i}\}_{i=1}^{T}$ and each
ingredient contains $L$ tokens $x_{i} = \{x_{ij}\}_{j=1}^{L}$. The
corresponding recipe is $y = \{y_{v}\}_{v=1}^{K}$. We would like to model
$p(y|X) = \Pi_v p(y_v|X, y_{<v})$. 

We first use a LSTM~\cite{hochreiter1997long} to encode each ingredient:
$h_{i, j} = \text{LSTM}_{\text{E}}(W_{E}x_{ij}, h_{i, j-1}) \quad \forall i$.
Then, we sum the resulting final state of each ingredient to obtain the
starting LSTM state of the decoder. We use an attention based
decoder:
\begin{align*}
  s_{v} & = \text{LSTM}_{\text{D}}([W_{\text{E}}y_{v-1}, d_{v-1}], s_{v-1}), \\
  p^{\text{copy}}_{v} & = \text{ATTN}(\{\{h_{i, j}\}_{i=1}^{T}\}^{L}_{j=1}, s_{v}) , \\
  d_{v} &= \sum_{i j}p_{v, i, j}h_{i, j}, \\
  p(z_{v}|s_{v}) &= \text{sigmoid}(W[s_{v}, d_{v}]), \\
  p_{v}^{\text{vocab}} & = \text{softmax}(W[s_{v},d_{v}]),
\end{align*}
where $\text{ATTN}(h, q)$ is the attention function that returns the
probability distribution over the set of vectors $h$, conditioned on any input
representation $q$. A full description of this operation is described in~\citep{BahdanauCB14}.
The decision to copy from the ingredient list or
generate a new word from the softmax is performed using a switch, denoted as
$p(z_{v}|s_{v})$. We can obtain a probability distribution of copying each of
the words in the ingredients by computing
$p^{\text{copy}}_{v} = \text{ATTN}(\{\{h_{i, j}\}_{i=1}^{T}\}^{L}_{j=1},
s_{v})$
in the attention mechanism.

\noindent {\bf Objective:} We can obtain the value of $z_{v}$ through a
string match of tokens in recipes with tokens in ingredients. If a token appears in the
ingredients, we set $z_{v}=1$ and $z_{v}=0$ otherwise.  We can train the model
in a fully supervised fashion, i.e., we can obtain the probability of $y_v$ as
$p(y_v, z_v| s_v) = p_v^{\text{copy}}(y_v) p(1|s_{v})$ if $z_{v}=1$ and
$p_v^{\text{vocab}}(y_v)(1- p(1|s_{i, v}))$ otherwise.

However, it may be not be accurate. In many cases, the tokens that appear in
the ingredients do not specifically refer to ingredients tokens. For examples, 
the recipe may start with ``Prepare a cup of water''. The token ``cup'' does not refer to the ``cup''
in the ingredient list ``1 cup plain soy milk''. To solve this problem, we treat $z_i$ as a 
latent variable, we wish to maximize the marginal probability of $y_{v}$ over 
all possible values of $z_v$. In this way, the model can automatically learn when to refer 
to tokens in the ingredients. Thus, the probability of generating token $y_v$ is defined as:
\begin{align*}
  p(y_{v} | s_{v}) &= p_v^{\text{vocab}}(y_v)p(0|s_{v})  +
                           p_v^{\text{copy}}(y_v)p(1|s_{v}) \\
                         & = p_v^{\text{vocab}}(y_v)(1- p(1|s_{v})) + p_v^{\text{copy}}(y_v)p(1|s_{v}).
\end{align*}
If no string match is found for $y_{v}$, we simply set $p_v^{\text{copy}}(y_v) = 0$
in the above objective.

\subsection{Reference to databases}
\begin{table*}[!thp]
\begin{small}
  \centering
  \begin{tabular}{| p{1.2cm} | l | l | l | p{3.3cm} | l | l |}
    \hline
    \textsc{name} & \textsc{price range} & \textsc{food} & \textsc{area} & \textsc{address} & \textsc{post code} & \textsc{phone} \\
    \hline
    ali baba & moderate & lebanese & centre & 59 Hills Road City Centre & CB 2, 1
                                                                          NT &
                                                                               01462 432565 \\
    \hline
    % anatolia & moderate & turkish & centre & 30 Bridge Street City Centre &
    % CB 2, 1 UJ &
    % 01223 362372 \\
    % \hline
    the nirala & moderate & indian & north & 7 Milton Road Chesterton & CB 4, 1 UY & 01223 360966 \\
    \hline
  \end{tabular}
  \caption{Fragment of database for dialogue system.}
  \label{tab:table}
  \end{small}
\end{table*}
We then consider the more complicated task of reference to database entries.
Referring to databases is quite common in question answering and dialogue systems,
in which databases are external knowledge and they are resorted to reply users'
query. In our paper, we consider the application of task-oriented dialogue systems 
in the domain of restaurant recommendations. Different from lists that are one dimensional,
databases are two-dimensional and referring to table entries requires sophisticated model 
design.

To better understand the model, we first make a brief introduction of the data set.
We use dialogues from the second Dialogue State Tracking Challenge
(DSTC2)~\citep{henderson2014dialog}. Table.~\ref{tab:example} is one example
dialogue from this dataset.
\begin{table}[!thbp]
\centering
\begin{tabular}{r  p{6cm}}
  M:& Hello, welcome to the Cambridge restaurant system? You can ask for
      restaurants by \texttt{area}, \texttt{price range} or \texttt{food type} . How may I help you?  \\
  U:& i want a \texttt{moderately priced} restaurant in the \texttt{north} town \\
  M:& \texttt{the nirala} is a nice restaurant in the north of town in the moderate price
      range \\
  U:& what is the \texttt{address} \\
  M:& Sure , \texttt{the nirala} is on \texttt{7 Milton Road Chesterton} \\
%  U:& what is the \texttt{phone number} \\
 % M:& The phone number of \texttt{the nirala} is \texttt{01223 360966} . \\
 % U:& thank you good bye \\
\end{tabular}
\caption{Example dialogue, M stands for Machine and U stands for User}
\label{tab:example}
\end{table}

We can observe from this example, users get recommendations of restaurants
based on queries that specify the area, price and food type of the
restaurant. We can support the system's decisions by incorporating a mechanism
that allows the model to query the database to find
restaurants that satisfy the users' queries.
% Thus, we crawled TripAdvisor for
% restaurants in the Cambridge area, where the dialog dataset was
% collected. Then, we remove restaurants that do not appear in the data set and
% create a database with 109 entries with restaurants and their attributes
% (e.g. food type).
A sample of our database (refer to data preparation part on how we construct
the database) is shown in Table~\ref{tab:table}. We can observe that each
restaurant contains 6 attributes that are generally referred in the dialogue
dataset. As such, if the user requests a restaurant that serves ``indian''
food, we wish to train a model that can search for entries whose ``food''
column contains ``indian''. Now, we describe how we deploy a model that
fulfills these requirements. We first introduce the basic dialogue framework
in which we incorporates the table reference module.
%Traditional dialogue systems are divided into several parts, including a
%natural language understanding unit, a state tracking unit, a table query unit
%and a natural language generator unit. Most of them rely on template-based
%hand-crafted rules. These templates and rules require domain specific efforts
%and do not generalize to other domains. We propose an end-to-end neural
%dialogue model which learns to query the table automatically. Our model is
%based on the recent success of sequence to sequence
%learning~\citep{sutskever2014sequence}.
%We augment the sequence generation model with a table query part and our model
%automatically learns to decide when to generate tokens from table and if yes,
%which token to copy from table. We begin by introducing our baseline model
%which does not condition on the table.

\begin{figure}[!t]
  \centering
  \includegraphics[width=0.5\textwidth]{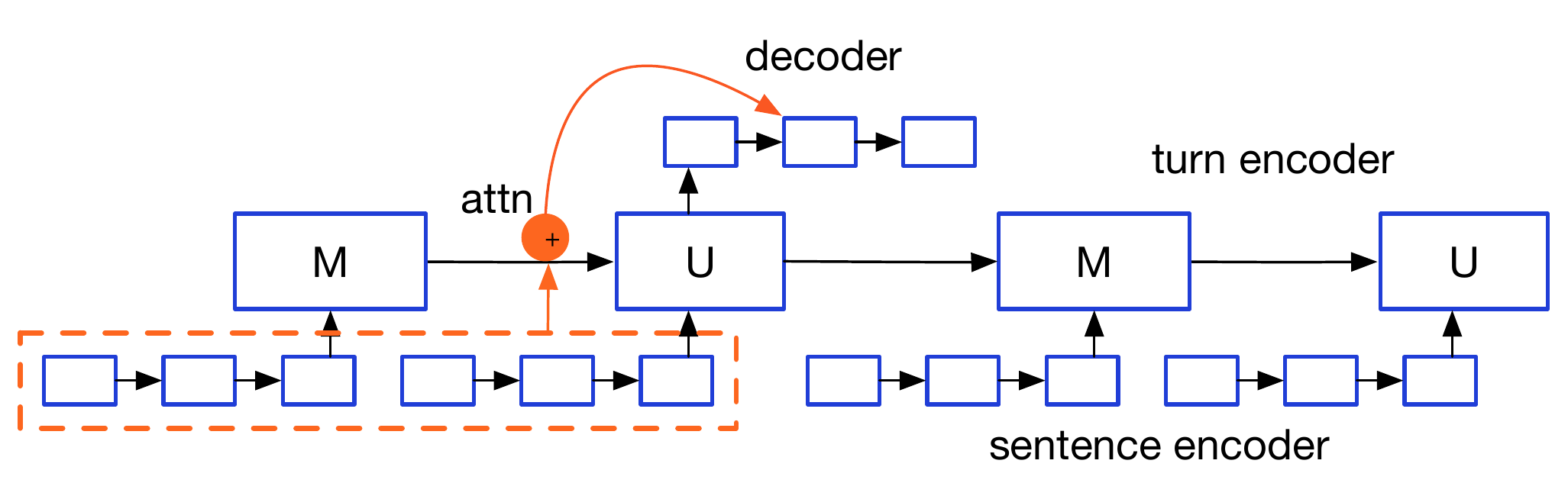}
  \caption{Hierarchical RNN Seq2Seq model. The red box denotes attention
  mechanism over the utterances in the previous turn.}
  \label{fig:hseq2seq}
  \vspace{-0.3cm}
\end{figure}
\noindent\textbf{Basic Dialogue Framework:} We build a basic dialogue model based on the hierarchical RNN model described
in~\citep{serban2016building}, as in dialogues, the generation of the response
is not only dependent on the previous sentence, but on all sentences leading to
the response.  We assume that a dialogue is alternated between a machine and a
user. An illustration of the model is shown in Figure~{\ref{fig:hseq2seq}}.

Consider a dialogue with $T$ turns, the utterances from a user and a machines
are denoted as $X =\{x_{i}\}^{T}_{i=1} $ and $Y=\{y_{i}\}_{i=1}^{T}$
respectively, where $i$ is the $i$-th utterance. We define
$x_{i} = \{x_{ij}\}_{j=1}^{|x_{i}|}$, $y_{i} = \{y_{iv}\}_{v=1}^{|y_{i}|}$,
where $x_{ij}$ ($y_{iv}$) denotes the $j$-th ($v$-th) token in the $i$-th
utterance from the user (the machines). The dialogue sequence starts with a
machine utterance and is given by
$\{y_{1}, x_{1}, y_{2}, x_{2}, \ldots , y_{T}, x_{T}\}$. We would like to model
the utterances from the machine
\begin{align*}
  & p(y_{1}, y_{2}, \ldots ,y_{T}| x_{1}, x_{2}, \ldots , x_{T}) = \\
  & \quad \prod_{i} p(y_{i} |
    y_{<i}, x_{<i}) = \prod_{i,v} p(y_{i, v} | y_{i, <v}, y_{<i}, x_{<i}).
\end{align*}
% A neural model is employed to predict
% $p(y_{i, v} | y_{i, <v}, y_{<i}, x_{<i})$, which operates as follows:

We encode $y_{<i}$ and $x_{<i}$ into continuous space in a hierarchical way
with LSTM: {\bf Sentence Encoder}: For a given utterance $x_{i}$, 
%we start with an initial state $h^{x}_{i,0}$ and apply the recursion
We encode it as $h^{x}_{i, j} = \text{LSTM}_{\text{E}}(W_{E}x_{i, j}, h^{x}_{i, j-1})$.
% where $W_{E}x_{i,j}$ denotes a word embedding lookup for the token $x_{i,j}$. 
The representation of $x_{i}$ is given by the $h^{x}_{i} = h^{x}_{i, |x_{i}|}$.
The same process is applied to obtain the machine utterance representation
$h^{y}_{i} = h^{y}_{i, |y_{i}|}$. {\bf Turn Encoder}: We further encode the sequence
$\{h^{y}_{1}, h^{x}_{1},..., h^{y}_{i}, h^{x}_{i}\}$ with another LSTM
encoder. 
% Once again, we start with an initial state $u_0$ and feed each of the
% utterance representation to obtain the following state, until the final state
% is obtained. 
% For simplicity, 
We shall refer the last hidden state as $u_{i}$, which can be
seen as the hierarchical encoding of the previous $i$ utterances.
%sequence $\{h^{y}_{1}, h^{x}_{1},..., h^{y}_{T}, h^{x}_{T}\}$:
%$u^{x}_{i} = \text{LSTM}_{\text{TE}}(h^{x}_{i}, u^{y}_{i}), u^{y}_{i} =
%\text{LSTM}_{\text{TE}}(h^{y}_{i}, u^{x}_{i-1})$
{\bf Decoder}: 
% As for decoding, in order to generate each utterance
% $y_{i}$, we can feed $u_{i-1}$ into the decoder LSTM as the initial state,
We use $u_{i-1}$ as the initial state of decoder LSTM and
decode each token in $y_{i}$. We can express the decoder as:
\begin{align*}
  s^{y}_{i, v} & = \text{LSTM}_{\text{D}}(W_{E}y_{i, v-1}, s_{i, v-1}), \\
  p^{y}_{i, v} & = \text{softmax}(Ws^{y}_{i, v}).
\end{align*}
% where the desired probability $p(y_{i, v} | y_{i, <v}, y_{<i}, x_{<i})$ is
% denoted by $p^{y}_{i, v}$.
% \\[0.15cm]

We can also incoroprate the attetionn mechanism in the decoder. 
As shown in Figure.~\ref{fig:hseq2seq}, we use the attention mechanism 
over the utterance in the previous turn. Due to space limit, 
we don't present the attention based decoder mathmatically and readers can refer
to~\cite{BahdanauCB14} for details.

\subsubsection{Incorporating Table Reference}
We now extend the decoder in order to allow the model to condition 
the generation on a database. 

\noindent {\bf Pointer Switch}: We use $z_{i, v}\in \{0, 1\}$ to denote the
decision of whether to copy one cell from the table. We compute this
probability as follows:
\begin{align*}
  p(z_{i, v}|s_{i, v}) &= \text{sigmoid}(Ws_{i, v}).
\end{align*}
Thus, if $z_{i, v} = 1$, the next token $y_{i,v}$ is generated from the
database, whereas if $z_{i, v} = 0$, then the following token is generated from
a softmax. We now describe how we generate tokens from the database.

\begin{figure}[!tb]
%   \begin{subfigure}[t]{\linewidth}
%     \includegraphics[width=\linewidth]{table_attn.pdf}
%     \centering
%     \caption{Decoder with table attention.}
%     \label{fig:table_attn}
%   \end{subfigure}
%   \begin{subfigure}[t]{\linewidth}
    \includegraphics[width=\linewidth]{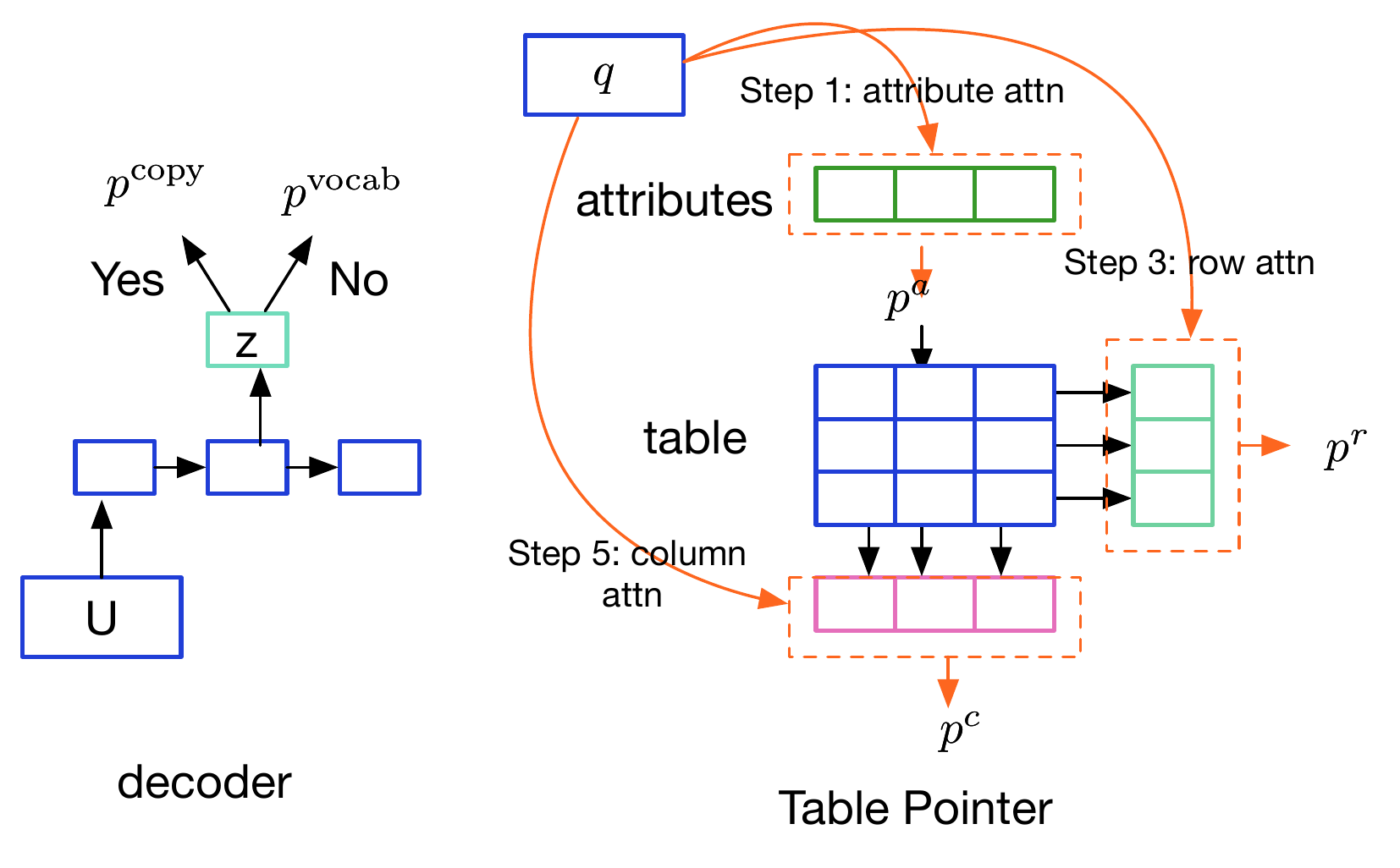}
    \centering
    \vspace{-0.4cm}
    \caption{Decoder with table pointer.}
    \label{fig:table_pointer}
%   \end{subfigure}
%   \caption{Table based decoder.}
\vspace{-0.4cm}
\end{figure}

We denote a table with $R$ rows and $C$ columns as
$\{t_{r,c}\}, r \in [1, R], c \in [1, C]$, where $t_{r,c}$ is the cell in row
$r$ and column $c$. The attribute of each column is denoted as $s_c$, where $c$
is the $c$-th attribute. $t_{r, c}$ and $s_{c}$ are one-hot vector.
\\[0.15cm]
\noindent {\bf Table Encoding}: To encode the table, we first build an
attribute vector and then an encoding vector for each cell. The attribute
vector is simply an embedding lookup $g_{c} = W_{E}s_{c}$. For the encoding of
each cell, we first concatenate embedding lookup of the cell with the
corresponding attribute vector $g_c$ and then feed it through a one-layer MLP
as follows: then $e_{r,c} = \tanh(W [W_{E}t_{r,c}, g_{c}])$.
\\[0.15cm]
\noindent {\bf Table Pointer}: The detailed process of calculating the
probability distribution over the table is shown in
Figure~\ref{fig:table_pointer}. The attention over cells in the table is
conditioned on a given vector $q$, similarly to the attention model for
sequences. However, rather than a sequence of vectors, we now operate over a
table.

\noindent {\bf Step 1}: Attention over the attributes to find out the
attributes that a user asks about, $p^{a} = \text{ATTN}(\{g_{c}\}, q)$. Suppose
a user says \texttt{cheap}, then we should focus on the \texttt{price}
attribute.

\noindent {\bf Step 2}: Conditional row representation calculation,
$e_{r} = \sum_{c}p^{a}_{c}e_{r,c} \quad \forall r$. So that $e_{r}$ contains the
price information of the restaurant in row $r$.

\noindent {\bf Step 3}: Attention over $e_{r}$ to find out the restaurants that
satisfy users' query, $p^{r} = \text{ATTN}(\{e_r\}, q)$. Restaurants with
\texttt{cheap} price will be picked.

\noindent {\bf Step 4}: Using the probabilities $p^{r}$, we compute the
weighted average over the all rows $e_{c} = \sum_{r} p^{r}_{r} e_{r,c}$.
$\{e_r\}$ contains the information of \texttt{cheap} restaurant.

\noindent {\bf Step 5:} Attention over columns $\{e_{r}\}$ to compute the
probabilities of copying each column $p^{c} = \text{ATTN}(\{e_{c}\}, q)$.

\noindent {\bf Step 6}: To get the probability matrix of copying each cell, we
simply compute the outer product $p^{\text{copy}} = p^{r}\otimes p^{c}$.

\noindent The overall process is as follows:
\begin{align*}
  p^{a} & = \text{ATTN}(\{g_{c}\}, q), \\
  e_{r} &= \sum_{c}p^{a}_{c}e_{r,c} \quad \forall r, \\
  p^{r} &= \text{ATTN}(\{e_{r}\}, q), \\
  e_{c} &= \sum_{r} p^{r}_{r}e_{r,c} \quad \forall c, \\
  p^{c} &= \text{ATTN}(\{e_{c}\}, q), \\
  p^{\text{copy}} &= p^{r}\otimes p^{c}.
\end{align*}

If $z_{i,v} = 1$, we embed the above attention process in the decoder by replacing the
conditioned state $q$ with the current decoder state $s^{y}_{i, v}$. 

% Although the above table attention mechanism can condition on table entries in
% text generation, it is a soft decision that does not model reference
% explicitly. We argue that modeling reference in databases explicitly through a
% hard decision makes it easier to learn reasoning over databases. We now
% describe the mechanism used to refer to specific database entries during
% decoding. At each time step, the model needs to decide whether to generate the
% next token from an entry of the database or from the word softmax. This is
% performed as follows.
% \\[0.15cm]

\noindent {\bf Objective:} As in previous task, we can train the model
in a fully supervised fashion, or we can treat the decision as a latent variable.
We can get $p(y_{i,v}|s_{i,v})$ in a similar way.

% i.e., we simply maximize
% $p^{\text{copy}}p(1|s_{i, v})$ if $z_{i,v}=1$ and
% $p^{\text{vocab}}(1- p(1|s_{i, v}))$ otherwise.
% \begin{align*}
%   p(y_{i, v} | s_{i, v}) &= p^{\text{vocab}}p(0|s_{i, v})  +
%                            p^{\text{copy}}p(1|s_{i, v}) \\
%                          & = p^{\text{vocab}}(1- p(1|s_{i, v})) + p^{\text{copy}}p(1|s_{i, v}).
% \end{align*}

%We can provide supervision on $z_{i, v}$ in the training. We do a string matching
%of token $y_{i, v}$, if it appears in the table, then we assume $z_{i, v} = 1$,
%otherwise $z_{i, v} = 0$, then the objective function is:
%\begin{align*}
%  p(y_{i, v} | s_{i, v}) = p(y_{i,v}| z_{i, v}, s_{i, v}) p(z_{i,
%  v} | s_{i, v}).
%\end{align*}
%Hence
%\begin{equation*}
%  p(y_{i, v} | s_{i, v}) =
%  \begin{cases}
%    p_{\text{vocab}}p(0|s_{i, v}) & \text{if} \quad z_{i, v} = 0 \\
%    p_{\text{copy}}p(1|s_{i, v}) & \text{else} \quad z_{i, v} = 1
%  \end{cases}
%\end{equation*}

%We can also treat the decision as a latent variable, then
%\begin{align}
%  p(y_{i, v} | s_{i, v}) &= p_{\text{vocab}}p(0|s_{i, v})  +
%  p_{\text{copy}}p(1|s_{i, v}) \\
%  &= p_{\text{vocab}}(1- p(1|s_{i, v}))  + p_{\text{copy}}p(1|s_{i, v})
%\end{align}
%If $z_{i, v} = 1$, but the token is not in the table, then $p_{\text{copy}} = 0$.

\subsection{Reference to document context}

Finally, we address the references that happen in a document itself and build a
language model that uses coreference links to point to previous
entities. Before generating a word, we first make the decision on whether it is
an entity mention. If so, we decide which entity this mention belongs to, then
we generate the word based on that entity. Denote the document as
$X=\{x_{i}\}_{i=1}^{L}$, and the entities are $E=\{e_{i}\}_{i=1}^{N}$, each
entity has $M_{i}$ mentions, $e_{i} = \{m_{ij}\}_{j=1}^{M_{i}}$, such that
$\{x_{m_{ij}}\}_{j=1}^{M_{i}}$ refer to the same entity. We use a LSTM to model
the document, the hidden state of each token is
$h_{i} = \text{LSTM}(W_{E}x_{i}, h_{i-1})$. We use a set
$h^{e} = \{h^{e}_{0}, h^{e}_{1}, ..., h^{e}_{M}\}$ to keep track of the entity
states, where $h^{e}_{j}$ is the state of entity $j$.

\begin{figure*}[tbh]
  \centering
  \begin{minipage}[b]{0.7\textwidth}
    \centering
    \begin{tabular}{p{12cm}}
      um and [I]\textsubscript{1} think that is what’s - Go ahead
      [Linda]\textsubscript{2}. Well and thanks goes to [you]\textsubscript{1} and to
      [the media]\textsubscript{3} to help [us]\textsubscript{4}...So
      [our]\textsubscript{4} hat is off to all of [you]\textsubscript{5}...
    \end{tabular}
  \end{minipage}
  \centering
  \begin{minipage}[b]{0.7\textwidth}
    \centering
    \includegraphics[width=0.9\linewidth]{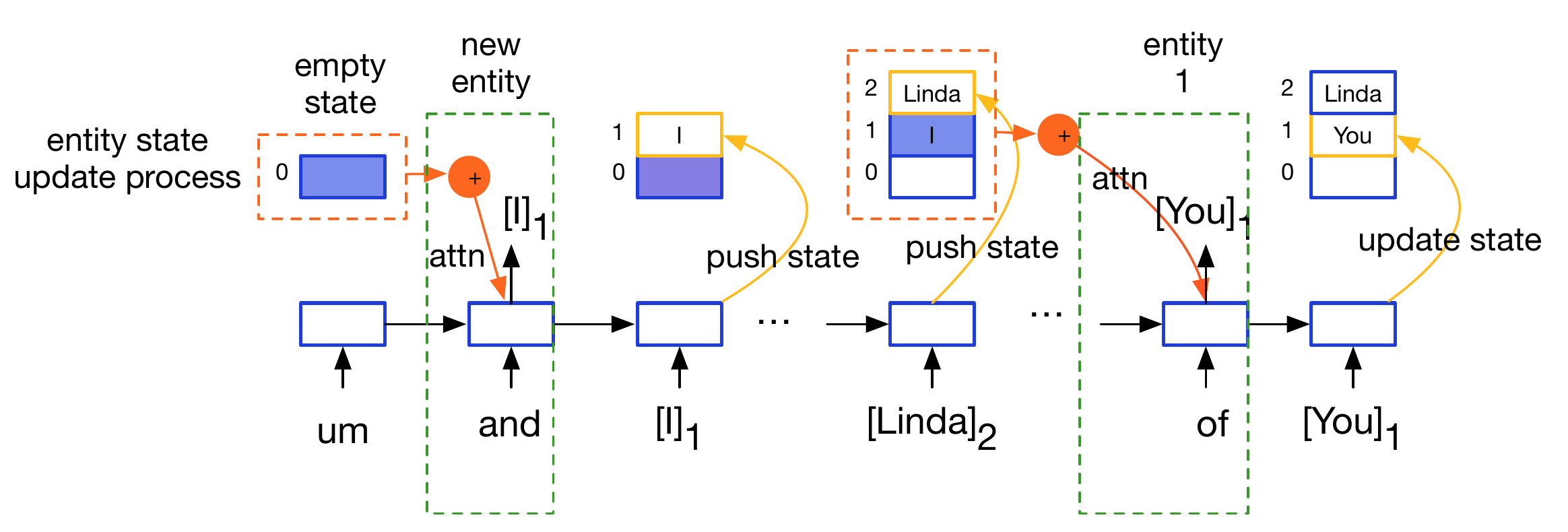}
  \end{minipage}
  \caption{Coreference based language model, example taken from \citet{wiseman2016learning}.}
  \label{fig:coref_lm}
  \vspace{-0.3cm}
\end{figure*}
\noindent {\bf Word generation}: At each time step before generating the next
word, we predict whether the word is an entity mention:
\begin{align*}
  p^{\text{coref}}(v_{i} | h_{i-1}, h^e)  & = \text{ATTN}(h^{e}, h_{i-1}), \\
  d_{i}  & = \sum_{v_{i}} p(v_{i}) h^{e}_{v_{i}},  \\
  p(z_{i} | h_{i-1})  & = \text{sigmoid}(W[h_{i-1}, d_{i}]),
\end{align*}
where $z_{i}$ denotes whether the next word is an entity and if yes $v_{i}$
denotes which entity the next word corefers to.  If the next word is an entity
mention, then
$p(x_{i} | v_{i}, h_{i-1}, h^{e}) =
\text{softmax}(W_{1}\tanh(W_{2}[h_{i-1}, h_{v_{i}}^{e}]))$
else $p(x_{i} | h_{i-1}) = \text{softmax}(W_{1} h_{i-1})$. Hence,
\begin{align*}
  & p(x_{i} | x_{<i}) = \\
  & \begin{cases}
    p(x_{i}| h_{i-1})p(z_{i} | h_{i-1}, h^{e}) & \enskip \text{if}\enskip z_{i} = 0. \\
    p(x_{i}|v_{i}, h_{i-1}, h^{e})\times\\\enskip p^\text{coref}(v_{i} |h_{i-1}, h^{e})p(z_{i}
    | h_{i-1}, h^{e}) & \enskip \text{if} \enskip z_{i} = 1.
  \end{cases}
\end{align*}
{\bf Entity state update}: Since there are multiple mentions for each entity
and the mentions appear dynamically, we need to keep track of the entity state
in order to use coreference information in entity mention prediction. We update the entity state $h^{e}$ at each
time step. In the beginning, $h^{e} = \{h^{e}_{0}\}$, $h^{e}_{0}$ denotes the
state of an virtual empty entity and is a learnable variable.
If $z_{i}=1$ and $v_{i} = 0$, then
it indicates the next word is a new entity mention, then in the next step,  we
append $h_{i}$ to $h^{e}$, i.e., $h^{e} = \{h^{e}, h_{i}\}$, if $z_{i} = 1$ and $v_{i} >
0$, then we update the corresponding entity state with the new hidden state,
$h^{e}[v_{i}] = h_{i}$. Another way to update the entity state is to use
one LSTM to encode the mention states and get the new entity state. Here we use the latest entity
mention state as the new entity state for simplicity.
The detailed update process is shown in Figure~\ref{fig:coref_lm}.

Note that the stochastic decisions in this task are more complicated than
previous two tasks. We need to make two sequential decisions: whether the next
word is an entity mention, and if yes, which entity the mention corefers to. It
is intractable to marginalize these decisions, so we train this model in a
supervised fashion (refer to data preparation part on how we get coreference
annotations).

\section{Experiments}
% \subsection{Data sets}
% For the dialogue modeling task, we use the DSTC2 dataset. For recipe
% generation, we crawled all recipes from \url{www.allrecipes.com}, and for the
% Coref LM, we use the Xinhua News portion of the Gigaword~v5 corpus. Details are
% included in Appendix A.
\subsection{Data sets and preprocessing}
{\bf Recipes}: We crawled all recipes from
\url{www.allrecipes.com}. There are about $31,000$ recipes in total, and every
recipe has an ingredient list and a corresponding recipe. We exclude the
recipes that have less than 10 tokens or more than 500 tokens, those recipes
take about 0.1\% of all data set.  On average each recipe has 118 tokens and 9
ingredients.  We random shuffle the whole data set and take 80\% as training
and 10\% for validation and test. We use a vocabulary size of 10,000 in the
model.

\noindent {\bf Dialogue}: We use the DSTC2 data
set. We only use the dialogue transcripts from the data set. There are about
3,200 dialogues in total. The table is not available from DSTC2.  To
reconstruct the table, we crawled TripAdvisor for restaurants in the Cambridge
area, where the dialog dataset was collected. Then, we remove restaurants that
do not appear in the data set and create a database with 109 restaurants and
their attributes (e.g. food type).  Since this is a small data set, we use
5-fold cross validation and report the average result over the 5
partitions. There may be multiple tokens in each table cell, for example in
Table.~\ref{tab:table}, the name, address, post code and phone number have
multiple tokens, we replace them with one special token. For the name, address,
post code and phone number of the $j$-th row, we replace the tokens in each
cell with $\_\text{NAME}\_j$, $\_\text{ADDR}\_j$, $\_\text{POSTCODE}\_j$,
$\_\text{PHONE}\_j$. If a table cell is empty, we replace it with an empty
token $\_\text{EMPTY}$. We do a string match in the transcript and replace the
corresponding tokens in transcripts from the table with the special
tokens. Each dialogue on average has 8 turns (16 sentences). We use a
vocabulary size of 900, including about 400 table tokens and 500 words.

\noindent {\bf Coref LM}: We use the Xinhua News data set from Gigaword Fifth
Edition and sample 100,000 documents that has length in range from 100 to
500. Each document has on average 234 tokens, so there are 23 million tokens in
total. We process the documents to get coreference annotations
and use the annotations, i.e., $z_{i}, v_{i}$, in training. We take 80\% as
training and 10\% as validation and test respectively. We ignore the entities
that have only one mention and for the mentions that have multiple tokens, we
take the token that is most frequent in the all the mentions for this
entity. After preprocessing, tokens that are entity mentions take about 10\% of
all tokens. We use a vocabulary size of 50,000 in the model.

\begin{table*}[!thbp]
\begin{small}
\centering
\begin{tabular}{l r r r r | r r r r}
  \toprule
  \multirow{3}{*}{Model} & \multicolumn{4}{c}{val} &  \multicolumn{4}{c}{test} \\
  \cline{2-9}
                         & \multicolumn{3}{c}{PPL} & \multirow{2}{*}{BLEU} & \multicolumn{3}{c}{PPL} &
                                                                                                       \multirow{2}{*}{BLEU} \\
  \cline{2-4} \cline{6-8}
                         & All & Ing & Word & & All & Ing & Word & \\
  \hline
  Seq2Seq & 5.60 & 11.26 & {\bf 5.00} & 14.07 & 5.52 & 11.26  & {\bf 4.91} & 14.39 \\
  Attn & 5.25 & 6.86 & 5.03 & 14.84 & 5.19 & 6.92 & 4.95 & 15.15 \\
  Pointer & 5.15 & 5.86 & 5.04 & {\bf 15.06} & 5.11 & 6.04 & 4.98 & 15.29 \\
  Latent & {\bf 5.02} & {\bf 5.10} & 5.01 & 14.87 & {\bf 4.97} & {\bf
                                                                 5.19} &

                                                                         4.94
                               & {\bf 15.41} \\
  \bottomrule
\end{tabular}
\caption{Recipe results, evaluated in perplexity and BLEU score. All means all tokens,
Ing denotes tokens from recipes that appear in ingredients. Word means non-table tokens.
Pointer and Latent differs in that for Pointer, we provide supervised signal on when 
to generate a reference token, while in Latent it is a latent decision.}
\label{tab:recipe}
\end{small}
\end{table*}

\begin{table*}[!thbp]
\begin{small}
\centering
\begin{tabular}{l r r r r}
  \toprule
  Model & All & Table & Table OOV & Word \\
  \midrule
  Seq2Seq & 1.35$\pm$0.01 & 4.98$\pm$0.38 & 1.99E7$\pm$7.75E6 & 1.23$\pm$0.01 \\
  Table Attn & 1.37$\pm$0.01 & 5.09$\pm$0.64 & 7.91E7$\pm$1.39E8 & 1.24$\pm$0.01 \\
  Table Pointer & {\bf 1.33$\pm$0.01} & {\bf 3.99$\pm$0.36} & {\bf 1360 $\pm$ 2600} & {\bf 1.23$\pm$0.01} \\
  Table Latent & 1.36$\pm$0.01 & 4.99$\pm$0.20 & 3.78E7$\pm$6.08E7 & 1.24$\pm$0.01 \\
  \midrule
  + {\bf Sentence Attn} \\
  Seq2Seq & 1.28$\pm$0.01 & 3.31$\pm$0.21 & 2.83E9 $\pm$ 4.69E9 & {\bf 1.19$\pm$0.01} \\
  Table Attn & 1.28$\pm$0.01 & 3.17$\pm$0.21 & 1.67E7$\pm$9.5E6 & 1.20$\pm$0.01 \\
  Table Pointer & {\bf 1.27$\pm$0.01} & {\bf 2.99$\pm$0.19} & {\bf 82.86$\pm$110} & 1.20$\pm$0.01 \\
  Table Latent & 1.28$\pm$0.01 & 3.26$\pm$0.25 & 1.27E7$\pm$1.41E7 & 1.20$\pm$0.01 \\
  \bottomrule
\end{tabular}
\caption{Dialogue perplexity results. Table means tokens
  from table, Table OOV denotes table tokens that do not appear in the
  training set. {\bf Sentence Attn} denotes we
  use attention mechanism over tokens in utterances from the previous
  turn. }
\label{tab:dialogue}
\vspace{-0.3cm}
\end{small}
\end{table*}

\subsection{Baselines, model training and evaluation}
We compare our model with baselines that do not model reference explicitly. 
For recipe generation and dialogue modeling, we compare our model with basic seq2seq and 
attention model. We also apply attention mechanism over the table for dialogue modeling as a baseline.
For coreference based language model, we compare our model with
simple RNN language model.

We train all models with simple stochastic gradient descent with gradient
clipping. We use a one-layer LSTM for all RNN components. Hyper-parameters are
selected using grid search based on the validation set.
% We use dropout after the input
% embedding and LSTM output. The learning rate is selected from [0.1, 0.2, 0.5,
% 1], maximum gradient norm is selected from [1, 2, 5, 10] and drop ratio is
% selected from [0.2, 0.3, 0.5]. The batch size and LSTM dimension size is
% slightly different for different tasks so as to make the model fit into
% memory. The number of epochs to train are different for each task and we drop
% the learning rate after reaching a given number of epochs.

Evaluation of our model is challenging since it involves three rather different
applications. We focus on evaluating the accuracy of predicting the reference
tokens, which is the goal of our model.  Specifically, we report the perplexity
of all words, words that can be generated from reference and non-reference
words. The perplexity is calculated by multiplying the probability of decision
at each step all together. Note that for non-reference words, they also appear
in the vocabulary. So it is a fair comparison to models that do not model
reference explicitly. For the recipe task, we also generate the recipes using
beam size of 10 and evaluate the generated recipes with BLEU. We didn't use
BLEU for dialogue generation since the database entries take only a very small
part of all tokens in utterances.

\subsection{Results and analysis}
The results for recipe generation, dialogue and coref based language model are shown
in Table~\ref{tab:recipe}, \ref{tab:dialogue}, and \ref{tab:coref_lm}
respectively. The recipe results in Table~\ref{tab:recipe} verifies that
modeling reference explicitly improves performance. Latent and Pointer perform
better than Seq2Seq and Attn model. The Latent model performs better than the Pointer model
since tokens in ingredients that match with recipes do not necessarily
come from the ingredients. Imposing a supervised signal gives wrong information
to the model and hence makes the result worse. With latent decision, the model
learns to when to copy and when to generate it from the vocabulary.

The findings for dialogue basically follow that of recipe generation, as shown in Table~\ref{tab:dialogue}. 
Conditioning table performs better in predicting table 
tokens in general. Table Pointer has the lowest perplexity for tokens in the table. Since the table
tokens appear rarely in the dialogue transcripts, the overall perplexity does
not differ much and the non-table token perplexity are similar. With attention
mechanism over the table, the perplexity of table token improves over basic
Seq2Seq model, but still not as good as directly pointing to cells in the table,
which shows the advantage of modeling reference explicitly. As expected, using
sentence attention improves significantly over models without sentence
attention. Surprisingly, Table Latent performs much worse than Table
Pointer. We also measure the perplexity of table tokens that appear only in
test set. For models other than Table Pointer, because the tokens never appear
in the training set, the perplexity is quite high, while Table Pointer can predict
these tokens much more accurately. This verifies our conjecture that our model
can learn reasoning over databases.

The coref based LM results are shown in Table~\ref{tab:coref_lm}. We find that coref
based LM performs much better on the entity perplexity, but is a little
bit worse for non-entity words. We found it was an optimization problem and the
model was stuck in a local optimum. So we initialize the Pointer model with the
weights learned from LM, the Pointer model performs better than LM both for
entity perplexity and non-entity words perplexity.

In Appendix A, we also visualize the heat map of table reference and list items
reference. The visualization shows that our model can correctly predict when to
refer to which entries according to context.

\begin{table}[!ht]
\centering
\begin{small}
\begin{tabular}{l r r r | r r r}
  \toprule
  \multirow{2}{*}{Model} & \multicolumn{3}{c}{val} &  \multicolumn{3}{c}{test} \\
                         & All & Entity & Word & All & Entity & Word \\
  \hline
  LM & 33.08 & 44.52 & 32.04 & 33.08 & 43.86 & 32.10 \\
  Pointer & 32.57 & 32.07 & 32.62 & 32.62 & 32.07 & 32.69 \\
  \parbox{1cm}{Pointer + init} & {\bf 30.43} & {\bf 28.56} & {\bf 30.63}
                               & {\bf 30.42} & {\bf 28.56} & {\bf 30.66} \\ \bottomrule
\end{tabular}
\caption{Coreference based LM. Pointer + init means we initialize the model
  with the LM weights.}
\label{tab:coref_lm}
\end{small}
  \vspace{-0.5cm}
\end{table}

%Endless amounts of dialog work, referring expression generation work, blah blah
%blah. Neural checklist models~\citep{kiddon:2016}

%Dialogue conditioning on semantics~\citep{wen:2016}, neural knowledge language
%model~\citep{ahn:2016}.

%Chit-chat dialogue systems as (conditional) language
%models~\citep{vinyals:2015,sordoni:2015,li:2016}.

\section{Related Work}

In terms of methodology, our work is closely related to previous works that
incorporate copying mechanism with neural
models~\citep{GulcehreANZB16,GuLLL16,ling:2016,ptrnets}. Our models are similar
to models proposed in~\citep{ahn:2016, merity2016pointer}, where the generation
of each word can be conditioned on a particular entry in knowledge lists and
previous words. In our work, we describe a model with broader applications,
allowing us to condition, on databases, lists and dynamic lists.

In terms of applications, our work is related to chit-chat
dialogue~\citep{li:2016, vinyals:2015, sordoni:2015, serban2016building,
  shang2015neural} and task oriented dialogue~\citep{wen:2016,
  bordes2016learning, williams2016end, wen2016network}. Most of previous works
on task oriented dialogues embed the seq2seq model in traditional dialogue
systems, in which the table query part is not differentiable, while our model
queries the database directly. Recipe generation was proposed
in~\citep{kiddon:2016}. They use attention mechanism over the checklists,
whereas our work models explicit references to them. Context dependent language
models~\citep{mikolov2010recurrent,
  jozefowicz2016exploring,mikolov2010recurrent, ji2015document, wang2015larger}
are proposed to capture long term dependency of text. There are also lots of
works on coreference resolution~\citep{haghighi2010coreference,
  wiseman2016learning}. We are the first to combine coreference with language
modeling, to the best of our knowledge.

\section{Conclusion}
We introduce reference-aware language models which explicitly model the
decision of from where to generate the token at each step. Our model can also
learns the decision by treating it as a latent variable. We demonstrate on
three applications, table based dialogue modeling, recipe generation and coref
based LM, that our model performs better than attention based model, which does
not incorporate this decision explicitly. There are several directions to
explore further based on our framework. The current evaluation method is based
on perplexity and BLEU. In task oriented dialogues, we can also try human
evaluation to see if the model can reply users' query accurately. 
It is also interesting to use reinforcement learning to learn the actions in
each step in coref based LM.

\bibliography{biblio}
\bibliographystyle{emnlp_natbib}

\appendix
\onecolumn
\section{Visualization}
\begin{table*}[!thbp]
  \centering
  \footnotesize
  \begin{subtable}{\textwidth}
    \begin{tabular}{r p{12cm}}
      M: &  Hello , welcome to the Cambridge restaurant system ? You can ask for
           restaurants by area , price range or food type . How may I help you ? \\
      U: & and expensive restaurant that serves tuscan food \\
      M: & Sorry there is no tuscan restaurant in the expensive price range \\
      U: & french food \\
      M: & \texttt{cote} is a great restaurant serving french food and it is in the
           expensive price range \\
      U: & may i have the phone number \\
      M: & The phone number of cote is \texttt{01223 311053} . \\
      U: & thank you good bye
    \end{tabular}
    \caption{Dialogue script}
  \end{subtable}
  \begin{subtable}{\textwidth}
  \begin{tabular}{| p{2cm} | c | c | c | p{3cm} | c | c |}
    \hline
    \textsc{name} & \textsc{price range} & \textsc{food} & \textsc{area} & \textsc{address} & \textsc{post code} & \textsc{phone} \\
    \hline
    charlie chan & cheap & chinese & east & Regent Street City Centre & C.B 2, 1 D.B & 01223 361763 \\
    \hline
    chiquito restaurant bar & expensive & mexican & south & 2G Cambridge Leisure Park Cherry Hinton Road Cherry Hinton & C.B 1, 7 D.Y & 01223 400170 \\
    \hline
    city stop & expensive & food & north & Cambridge City Football Club Milton Road Chesterton & $\_$EMPTY & 01223 363270 \\
    \hline
    clowns cafe & expensive & italian & centre & $\_$EMPTY & C.B 1, 1 L.N & 01223 355711\\
    \hline
    cocum & expensive & indian & west & 71 Castle Street City Centre & C.B 3, 0 A.H & 01223 366668 \\
    \hline
\cellattn{53}{cote} & expensive & french & centre & Bridge Street City Centre & C.B 2, 1 U.F & 01223 311053 \\
    \hline
curry garden & expensive & indian & centre & 106 Regent Street City Centre & $\_$EMPTY & 01223 302330 \\
    \hline
curry king & expensive & indian & centre & 5 Jordans Yard Bridge Street City Centre & C.B 1, 2 B.D & 01223 324351 \\
    \hline
curry prince & moderate & indian & east & 451 Newmarket Road Fen Ditton & C.B 5, 8 J.J & 01223 566388 \\
    \hline
  \end{tabular}
  \caption{Attention heat map: \texttt{cote} is a great restaurant serving
    french food and it is in the expensive price range.}
\end{subtable}
  \begin{subtable}{\textwidth}
  \begin{tabular}{| p{2cm} | c | c | c | p{3cm} | c | c |}
    \hline
    \textsc{name} & \textsc{price range} & \textsc{food} & \textsc{area} & \textsc{address} & \textsc{post code} & \textsc{phone} \\
    \hline
    charlie chan & cheap & chinese & east & Regent Street City Centre & C.B 2, 1 D.B & 01223 361763 \\
    \hline
    chiquito restaurant bar & expensive & mexican & south & 2G Cambridge Leisure Park Cherry Hinton Road Cherry Hinton & C.B 1, 7 D.Y & 01223 400170 \\
    \hline
    city stop & expensive & food & north & Cambridge City Football Club Milton Road Chesterton & $\_$EMPTY & 01223 363270 \\
    \hline
    clowns cafe & expensive & italian & centre & $\_$EMPTY & C.B 1, 1 L.N & 01223 355711\\
    \hline
    cocum & expensive & indian & west & 71 Castle Street City Centre & C.B 3, 0 A.H & 01223 366668 \\
    \hline
    cote & expensive & french & centre & Bridge Street City Centre & C.B 2, 1 U.F & \cellattn{99}{01223 311053} \\
    \hline
    curry garden & expensive & indian & centre & 106 Regent Street City Centre & $\_$EMPTY & 01223 302330 \\
    \hline
    curry king & expensive & indian & centre & 5 Jordans Yard Bridge Street City Centre & C.B 1, 2 B.D & 01223 324351 \\
    \hline
    curry prince & moderate & indian & east & 451 Newmarket Road Fen Ditton & C.B 5, 8 J.J & 01223 566388 \\
    \hline
  \end{tabular}
  \caption{Attention heap map: The phone number of cote is \texttt{01223
      311053} . }
\end{subtable}
\caption{Dialogue visualization 1}
\label{tab:example_vis}
\end{table*}

\begin{table*}[!thbp]
  \centering
  \footnotesize
  \begin{subtable}{\textwidth}
    \begin{tabular}{r p{12cm}}
      M: & Hello , welcome to the Cambridge restaurant system ? You can ask for
           restaurants by area , price range or food type . How may I help you ? \\
      U: & cheap restaurant \\
      M: & What kind of food would you like ? \\
      U: & in the west part of town \\
      M: & \texttt{la margherita} is a nice restaurant in the west of town in the cheap price
           range \\
      U: & address \\
      M: & Sure , la margherita is on \texttt{15 Magdalene Street City Centre} \\
      U: & thank you \\
      M: & la margherita is a nice restaurant in the west of town in the cheap price
           range \\
      U: & good bye
    \end{tabular}
    \caption{Dialogue script}
  \end{subtable}
  \begin{subtable}{\textwidth}
  \begin{tabular}{| p{2cm} | c | c | c | p{3cm} | c | c |}
    \hline
    \textsc{name} & \textsc{price range} & \textsc{food} & \textsc{area} & \textsc{address} & \textsc{post code} & \textsc{phone} \\
    \hline
    india house & expensive & indian & west & 31 Newnham Road Newnham & $\_$EMPTY & 01223 461661 \\
    \hline
    j restaurant & cheap & oriental & centre & 86 Regent Street City Centre & C.B 2, 1 D.P & 01223 307581 \\
    \hline
    jinling noodle bar & moderate & chinese & centre & 11 Peas Hill City Centre & C.B 2, 3 P.P & 01223 566188 \\
    \hline
    kohinoor & cheap & indian & centre & 74 Mill Road City Centre & $\_$EMPTY & 01223 323639 \\
    \hline
    kymmoy & expensive & oriental & centre & 52 Mill Road City Centre & C.B 1, 2 A.S & 01223 311911 \\
    \hline
    \cellattn{44}{la margherita} & cheap & italian & west & 15 Magdalene Street City Centre & C.B 3, 0 A.F & 01223 315232 \\
    \hline
    la mimosa & expensive & mediterranean & centre & Thompsons Lane Fen Ditton & C.B 5, 8 A.Q & 01223 362525 \\
    \hline
    la raza & cheap & spanish & centre & 4 - 6 Rose Crescent & C.B 2, 3 L.L & 01223 464550 \\
    \hline
    la tasca & moderate & spanish & centre & 14 -16 Bridge Street & C.B 2, 1 U.F & 01223 464630 \\
    \hline
    lan hong house & moderate & chinese & centre & 12 Norfolk Street City Centre & $\_$EMPTY & 01223 350420 \\
    \hline
  \end{tabular}
    \caption{Attention heat map: \texttt{la margherita} is a nice restaurant in the west of town in the cheap price
           range}
  \end{subtable}
  \begin{subtable}{\textwidth}
  \begin{tabular}{| p{2cm} | c | c | c | p{3cm} | c | c |}
    \hline
    \textsc{name} & \textsc{price range} & \textsc{food} & \textsc{area} & \textsc{address} & \textsc{post code} & \textsc{phone} \\
    \hline
    india house & expensive & indian & west & 31 Newnham Road Newnham & $\_$EMPTY & 01223 461661 \\
    \hline
    j restaurant & cheap & oriental & centre & 86 Regent Street City Centre & C.B 2, 1 D.P & 01223 307581 \\
    \hline
    jinling noodle bar & moderate & chinese & centre & 11 Peas Hill City Centre & C.B 2, 3 P.P & 01223 566188 \\
    \hline
    kohinoor & cheap & indian & centre & 74 Mill Road City Centre & $\_$EMPTY & 01223 323639 \\
    \hline
    kymmoy & expensive & oriental & centre & 52 Mill Road City Centre & C.B 1, 2 A.S & 01223 311911 \\
    \hline
    la margherita & cheap & italian & west & \cellattn{99}{15 Magdalene Street City Centre} & C.B 3, 0 A.F & 01223 315232 \\
    \hline
    la mimosa & expensive & mediterranean & centre & Thompsons Lane Fen Ditton & C.B 5, 8 A.Q & 01223 362525 \\
    \hline
    la raza & cheap & spanish & centre & 4 - 6 Rose Crescent & C.B 2, 3 L.L & 01223 464550 \\
    \hline
    la tasca & moderate & spanish & centre & 14 -16 Bridge Street & C.B 2, 1 U.F & 01223 464630 \\
    \hline
    lan hong house & moderate & chinese & centre & 12 Norfolk Street City Centre & $\_$EMPTY & 01223 350420 \\
    \hline
  \end{tabular}
    \caption{Attention heap map: Sure , la margherita is on \texttt{15 Magdalene Street City Centre}. }
  \end{subtable}
  \caption{Dialogue visualization 2}
  \label{tab:example2}
\end{table*}

\begin{figure*}[tbh]
  \centering
  \begin{subfigure}[t]{1.0\linewidth}
    \includegraphics[width=\linewidth] {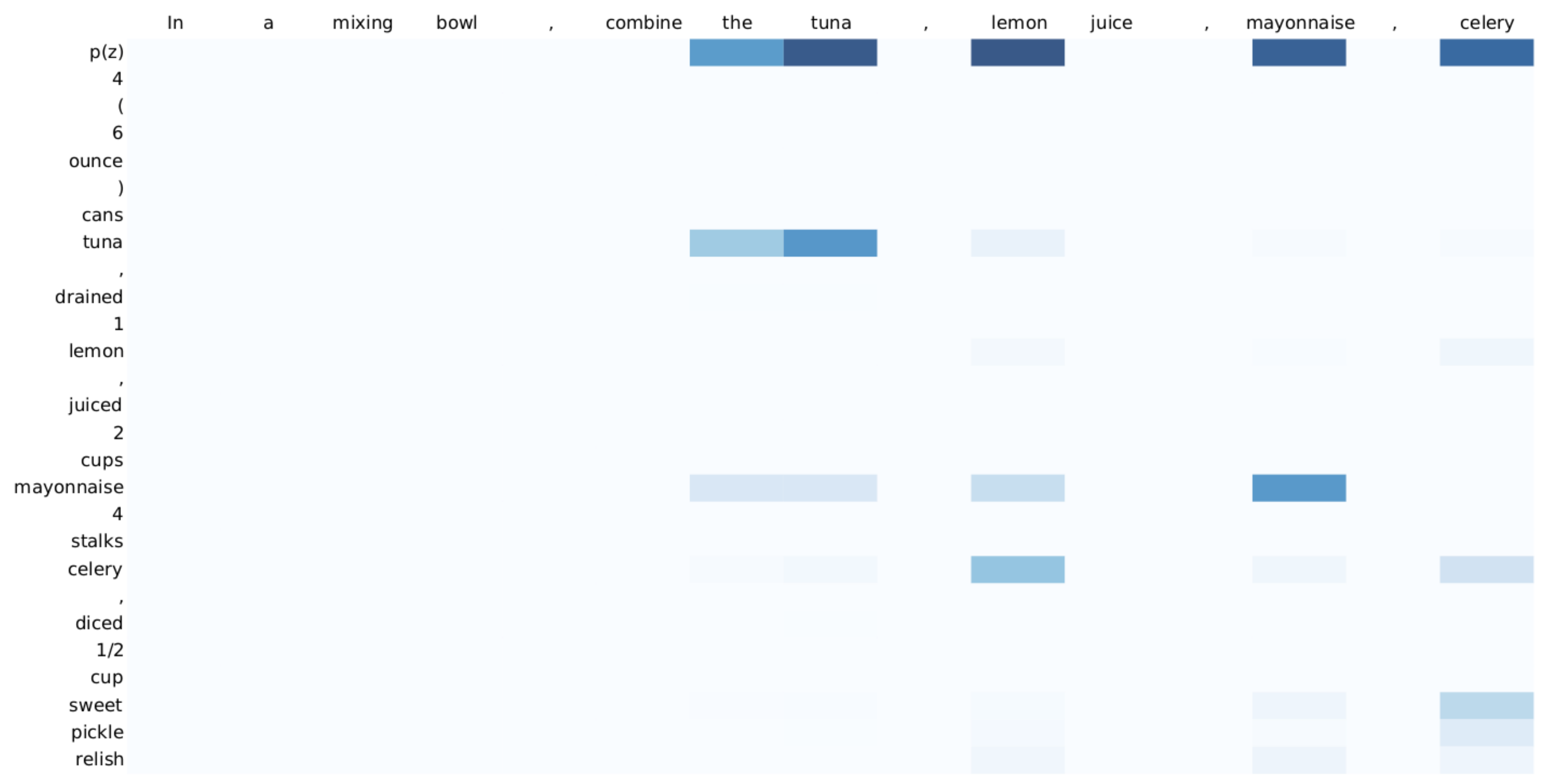}
    \centering
    \caption{part 1}
    \label{fig:recipe_map_0_0}
  \end{subfigure}

  \begin{subfigure}[t]{1.0\linewidth}
    \includegraphics[width=\linewidth] {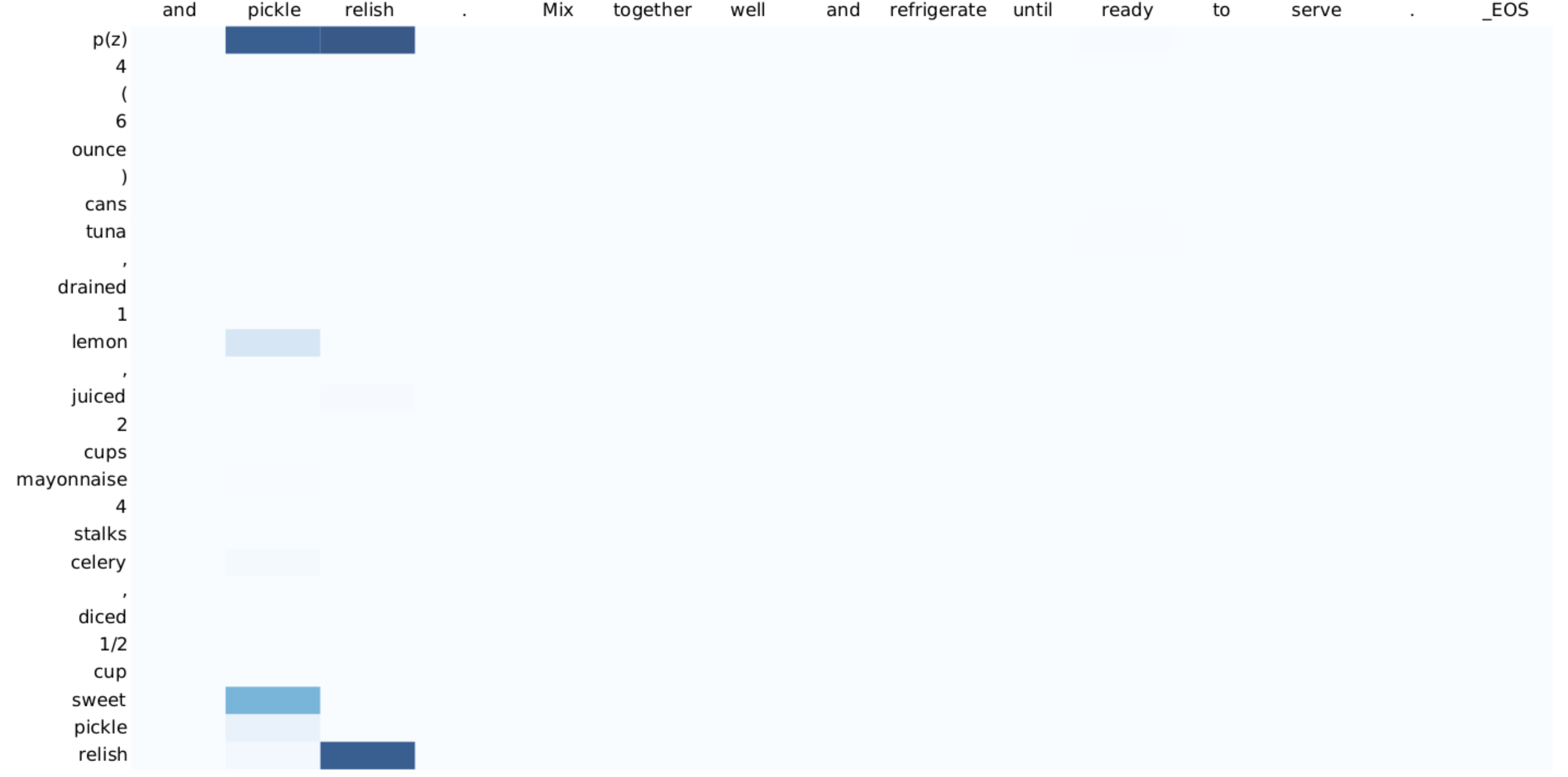}
    \centering
    \caption{part 2}
    \label{fig:recipe_map_0_1}
  \end{subfigure}
  \caption{Recipe heat map example 1. The ingredient tokens appear on the left
    while the recipe tokens appear on the top. The first row is the $p(z_{v}|s_{v})$.}
  \label{fig:recipe_map_0}
\end{figure*}

\begin{figure*}[tbh]
  \centering
  \begin{subfigure}[t]{1.0\linewidth}
    \includegraphics[width=\linewidth] {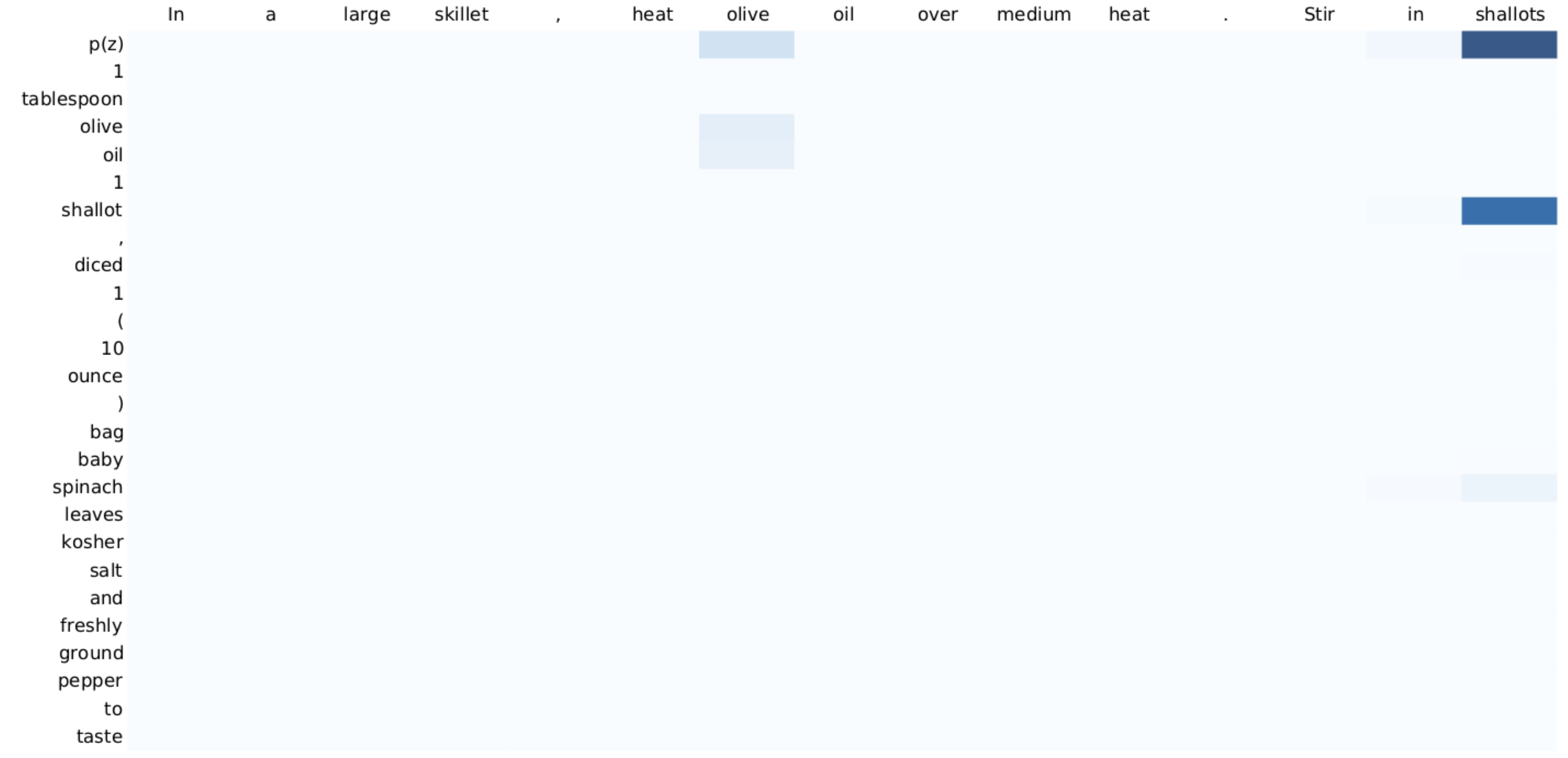}
    \centering
    \caption{part 1}
    \label{fig:recipe_map_1_0}
  \end{subfigure}

  \begin{subfigure}[t]{1.0\linewidth}
    \includegraphics[width=\linewidth] {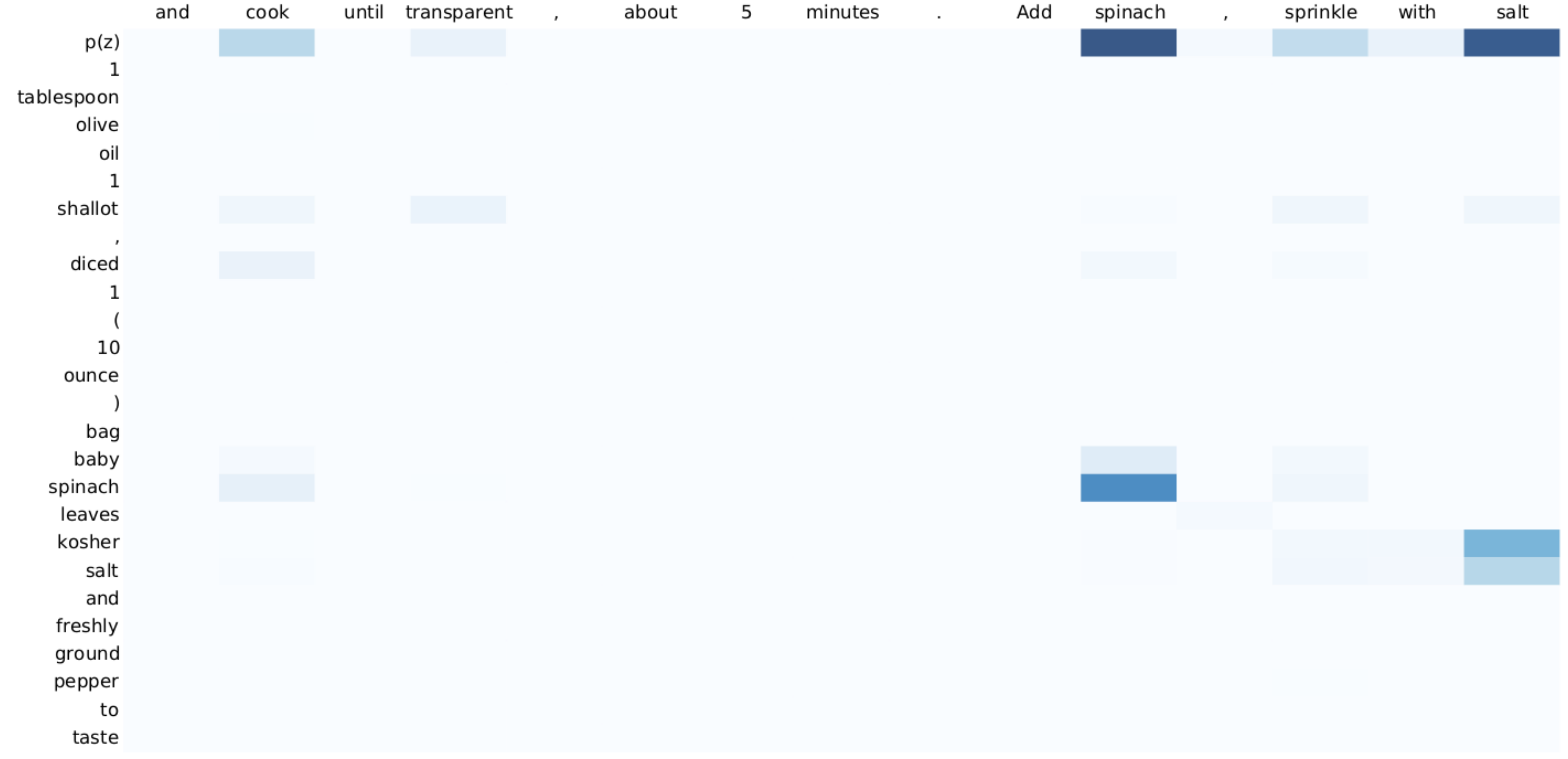}
    \centering
    \caption{part 2}
    \label{fig:recipe_map_1_1}
  \end{subfigure}
  \begin{subfigure}[t]{1.0\linewidth}
    \includegraphics[width=\linewidth] {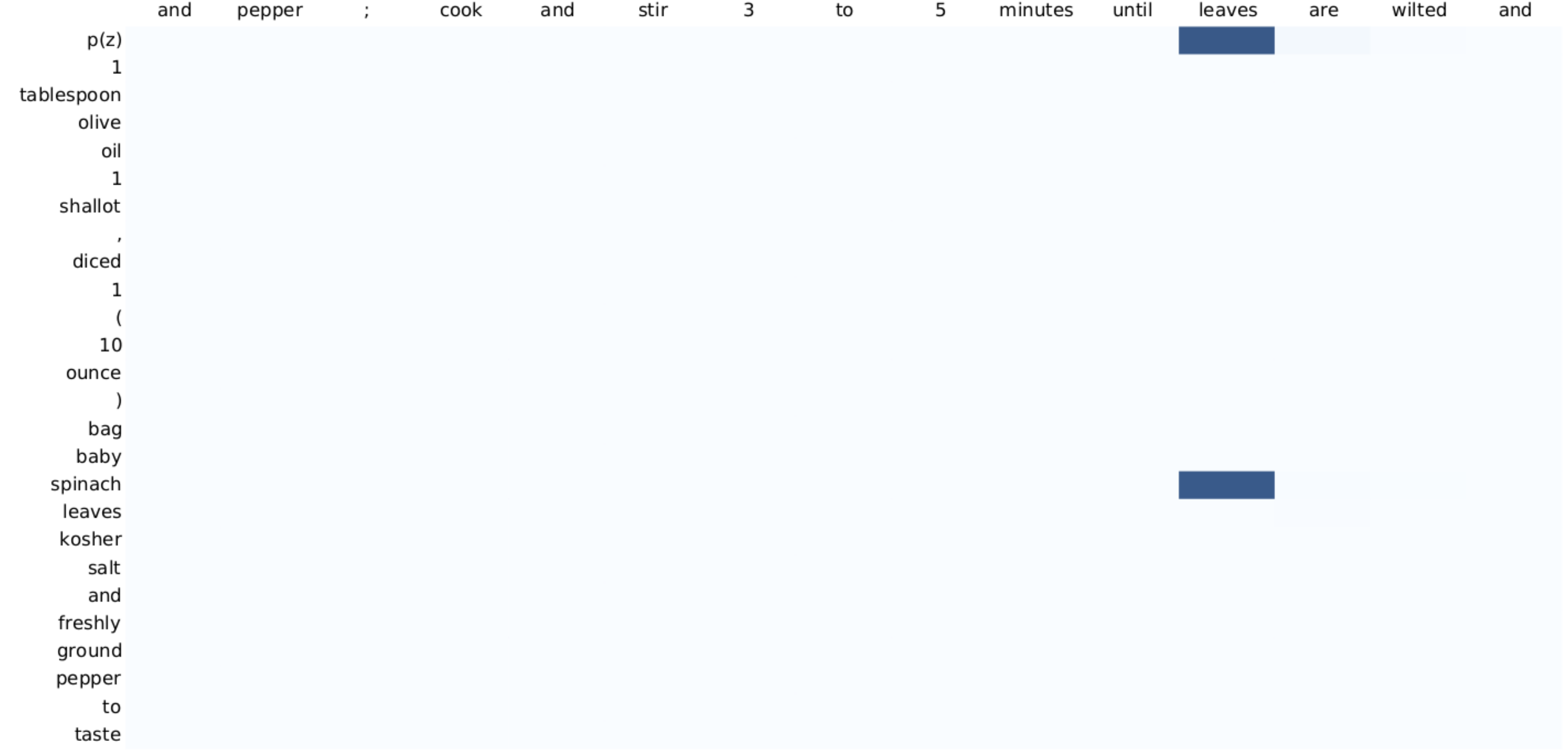}
    \centering
    \caption{part 3}
    \label{fig:recipe_map_1_2}
  \end{subfigure}
  \caption{Recipe heat map example 2.}
  \label{fig:recipe_map_1}
\end{figure*}

\end{document}